\pdfoutput=1

\documentclass[11pt]{article}

\usepackage[]{naacl2024-latex/NAACL2024}

\usepackage{times}
\usepackage{latexsym}
\usepackage{multirow}

\usepackage[T1]{fontenc}

\usepackage[utf8]{inputenc}

\usepackage{microtype}

\usepackage{inconsolata}

\usepackage{paralist}
\usepackage{booktabs}
\usepackage{graphicx}
\usepackage[export]{adjustbox}

\newcommand{\argsubg}[2]{[#1]\textsuperscript{\textsc{#2}}}

\usepackage{verbatim}

\pagecolor[rgb]{1,1,1}
\color[rgb]{0.1,0.1,0.1}

\title{Generating Uncontextualized and Contextualized Questions\\for Document-Level Event Argument Extraction}

\author{Md Nayem Uddin$^{\spadesuit}$\hspace{9pt}
Enfa Rose George$^{\diamondsuit}$ \hspace{9pt}
Eduardo Blanco$^{\diamondsuit}$ \hspace{9pt}
Steven R. Corman$^{\spadesuit}$ \vspace{9pt} \\
$^\spadesuit$Arizona State University \; $^\diamondsuit$ University of Arizona \vspace{2pt}\\
$^{\spadesuit}$\texttt{\{muddin11, steve.corman\}@asu.edu} \vspace{2pt}\\
$^{\diamondsuit}$\texttt{\{enfageorge, eduardoblanco\}@arizona.edu} \vspace{2pt}
}

\begin{document}
\maketitle
\begin{abstract} 
This paper presents multiple question generation strategies for document-level event argument extraction.
These strategies do not require human involvement and result in uncontextualized questions as well as contextualized questions grounded on the event and document of interest.
Experimental results show that combining uncontextualized and contextualized questions is beneficial,
especially when event triggers and arguments appear in different sentences.
Our approach does not have corpus-specific components, in particular, the question generation strategies transfer across corpora.
We also present a qualitative analysis of the most common errors made by our best model.\\
\end{abstract}

\section{Introduction}
\label{s:introduction}
Event argument extraction~\cite{doddington-etal-2004-automatic,aguilar-etal-2014-comparison} is about identifying entities participating
in events and specifying their role (e.g., the \emph{giver}, \emph{recipient}, and \emph{thing given} in a \emph{giving} event). 
Transforming natural language into structured event knowledge benefits
many downstream tasks such as
machine reading comprehension~\cite{han-etal-2021-ester}, news summarization~\cite{li-etal-2016-abstractive}, coreference resolution~\cite{huang-etal-2019-improving}, and dialogue systems~\cite{su-etal-2022-multi}. 

\begin{figure}[t!]
\centering
\input{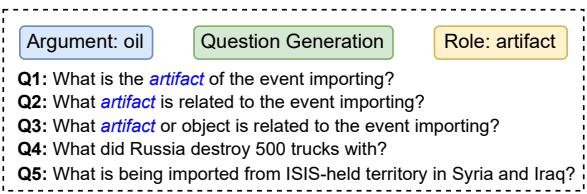}
\caption{
Example from RAMS (top, event trigger \emph{importing} and its arguments).
In this paper, we experiment with several strategies to generate questions for event argument extraction.
Questions for the \emph{artifact} of \emph{importing} are shown in the dashed box.
Q1 is generated following a role-specific template, 
Q2 and Q3 are generated prompting GPT-4,
and Q4 and Q5 are generated by a weakly-supervised T5 model.
}
\label{f:motivation}
\end{figure}

Traditionally, corpora 
are limited to arguments within the same sentence an event belongs to.
Inter-sentential arguments are more challenging
and have received less attention~\cite{gerber-chai-2010-beyond,ruppenhofer-etal-2010-semeval}.
Figure \ref{f:motivation} presents an example from RAMS~\cite{ebner-etal-2020-multi},
the largest corpus annotating multi-sentence event-argument structures.
Two out of four event-argument relations cross sentence boundaries. 

The current landscape of argument extraction is dominated 
by prompt-based methods~\cite{ma-etal-2022-prompt, nguyen-etal-2023-contextualized} 
and question-answering techniques~\citep{du-ji-2022-retrieval}. 
However, these approaches heavily rely on rigid templates, 
neglecting the valuable context provided by the document.
Another avenue for improvement involves integrating external knowledge~\cite{zhang-etal-2023-overlap,yang-etal-2023-amr}, 
but the challenge lies in the trial and error process
of finding relevant external knowledge. 
In this paper, we address these issues by generating contextualized questions grounded on the event of interest.
Our approaches capture the semantic relevance across 
different event arguments and non-event argument entities from the document.


As illustrated in Figure~\ref{f:motivation},
for the \emph{artifact} role of event trigger \emph{importing}, 
we generate five questions and employ transformers to identify answers
that pinpoint the argument (i.e., \emph{oil})
or, alternatively, indicate that there is no answer.
The first three questions are uncontextualized (i.e., disregard the document; they only take into account the event and argument role).
On the other hand, the last two questions are contextualized
(i.e., grounded on the event trigger and document).
Contextualized questions are generated with a T5 model trained using weak supervision.
As we shall see, they yield substantial improvements even though they are noisy (e.g., the answer to Q4 is not \emph{oil}).

The main contributions of this paper are:
\begin{compactitem}
\item Several strategies to generate uncontextualized and contextualized questions,
  including weakly supervised models that require no manual effort beyond defining prompts.
\item Experimental results showing that combining uncontextualized and contextualized questions is beneficial with RAMS.
\item Experimental results showing that our approach is effective with out-of-domain corpora, event triggers, and roles.
  In particular, the question generation component transfers seamlessly to other corpora (WikiEvents).
\end{compactitem}

A key benefit of our approach is that it can be easily adapted to any event-argument extraction benchmark---we do not have any benchmark-specific component.
The only requirement is
a list of argument roles an event may have.
Most event-argument annotation efforts satisfy this requirement,
including
PropBank~\cite{palmer-etal-2005-proposition},
NomBank~\cite{meyers-etal-2004-annotating}, 
FrameNet~\cite{baker-etal-1998-berkeley-framenet},
RAMS,
ACE~\cite{doddington-etal-2004-automatic},
and
WikiEvents~\cite{li-etal-2021-document}.
All examples in the paper are from RAMS, but we also applied our optimal question-answering strategy to WikiEvents to demonstrate its adaptability.

\section{Previous Work}
\label{s:previous_work}
Event argument extraction~\cite{ahn-2006-stages} has a long history in the field~\cite{grishman-sundheim-1996-message,doddington-etal-2004-automatic}.
Initially, datasets focused on extracting arguments within the same sentence than the event~\cite{palmer-etal-2005-proposition, walker2006ace}.
There are also corpora focused on inter-sentential arguments~\cite{gerber-chai-2010-beyond,ruppenhofer-etal-2010-semeval,ebner-etal-2020-multi,li-etal-2021-document}.
Early models were based on handcrafted features~\cite{li-etal-2013-joint,liao-grishman-2010-using,hong-etal-2011-using}.
Like most NLP tasks, models for event argument extraction
experienced a transformation building on word embeddings, RNNs, and CNNs~\cite{chen-etal-2015-event, nguyen-etal-2016-joint-event}.

Transformer-based approaches are currently the best performing.
Some efforts assume event triggers and argument spans are part of the input and present classifiers to identify the argument role~\cite{ebner-etal-2020-multi, chen-etal-2020-joint-modeling}.
Unlike them---and like the remaining previous works discussed below---we only assume event triggers.
At a high-level, efforts to identify argument spans and their role can be categorized into
sequence labeling, casting the problem as a question-answering task, and using generative models.
Sequence label classifiers approach the problem with the traditional BIO encoding~\cite{ramponi-etal-2020-biomedical}.
Framing the problem in terms of questions and answers is popular~\cite{du-cardie-2020-event, liu-etal-2020-event, li-etal-2020-event, uddin-etal-2024-eventqa}.
Doing so enables zero-shot~\cite{lyu-etal-2021-zero} and few-shot~\cite{sainz-etal-2022-textual} predictions.
\newcite{li-etal-2021-document} and \newcite{ma-etal-2022-prompt, du-etal-2021-grit} leverage generative language models~\cite{raffel2020exploring,lewis-etal-2020-bart}.
Language generation facilitates a more flexible extraction by \emph{generating} the arguments rather than identifying spans in the input document.
Transfer learning has also been explored, including
role overlapping knowledge~\cite{zhang-etal-2023-overlap},
semantic roles~\cite{zhang-etal-2022-transfer}, 
abstract meaning representations~\cite{xu-etal-2022-two}, 
and
frame-aware knowledge distillation~\cite{wei-etal-2021-trigger}.

We approach the problem as a question answering task by querying a transformer model.
Unlike previous work, we explore both uncontextualized as well as contextualized questions grounded on the event and document of interest.
The latter questions are generated with a weakly supervised model.
This allows us to train the model with one corpus~(RAMS)
that transfer to other corpus~(WikiEvents).

\section{Generating Questions for Event Argument Extraction}
\label{s:approaches}

We transform RAMS event-argument structures into questions and answers. 
While gold event-argument annotations include gold answers (i.e., the argument),
there are no gold questions asking for the arguments filling the roles of event triggers.
Hence, we propose five question generation strategies
divided into two categories: uncontextualized and contextualized.
Unlike uncontextualized questions, contextualized questions
are grounded on the event and document of interest.

\subsection{Uncontextualized Questions}
\label{ss:generating_independent_questions}
Uncontextualized questions only have access to the event trigger (e.g., \emph{importing} in Figure \ref{f:motivation})
and role~(e.g., \emph{artifact}).
They disregard the document of interest and always generate the same question for an event trigger and role.

\paragraph{Template-Based Questions.}
Inspired by previous work~\cite{du-cardie-2020-event, liu-etal-2021-machine},
we generate wh-questions for each argument role an event may have according to the event ontology. 
We use a straightforward template to generate questions: 
\emph{Wh-word is the [argument role] of event [event trigger]?}, 
where Wh-word is selected from the following: what, where, who and how. 
The only requirement is to know a priori 
the name of the argument roles an event may have,
an assumption we share with previous work.
Questions are generated regardless of whether the argument is present in the input document. 
Answers are \emph{No answer} if an argument is not present. 
For example, in Figure \ref{f:motivation}, “Q1: \emph{What is the artifact of the event importing?}” 
is the template-based question for the \emph{artifact} role of event trigger \emph{importing}.

\paragraph{Prompt-Based Questions.}
Template-based questions rely on templates defined by humans.
This is suboptimal when generating questions for new argument roles.
We employ large language models to bypass this limitation.
Specifically, we use zero- and few-shot prompting~(see Appendix~\ref{a:gpt4_question_generation} for details) 
with the GPT-4 model to generate wh-questions for all roles an event may have. 
The few-shot prompting includes 10 randomly chosen examples and enables in-context learning~\cite{brown2020language}.
These questions are designed to ask for specific roles
without being tied to any specific event document. 
We refer to these questions as zero- and few-shot prompt-based questions.
Q2 and Q3 in Figure \ref{f:motivation} are zero- and few-shot prompt-based questions generated for the \emph{artifact} of \emph{importing}:
\emph{What artifact is related to the event importing?} and 
\emph{What artifact or object is related to the event importing?} 

Uncontextualized questions sometimes result in unnatural or ungrammatical questions as exemplified below
(\emph{X} is a placeholder for the event trigger):
\begin{compactitem}
\item \emph{Where is the place of event [X]?} (template)
\item Zero Prompt: \emph{Where did the event [X] take place?} (zero-shot prompt)
\item Few Prompt: \emph{What is the place or location related to the event [X]?} (few-shot prompt)
\end{compactitem}

Rather than defining complex rules to account for the correct verb tense and formulate better wordings,
we work with contextualized questions.

\subsection{Contextualized Questions}
\label{ss:generating_dependent_questions}
Uncontextualized questions lack any information included in the document the event trigger belongs to,
including other arguments as well as the rest of the document.
Contextualized questions address this weakness.
We propose a weakly supervised method to generate contextualized questions grounded on the event and document of interest.
We explore two strategies to collect data to train models for question generation:
(a)~leveraging SQuAD~\citep{rajpurkar-etal-2016-squad}, a question-answering dataset not related to event argument extraction,
and
(b)~weak supervision from questions automatically obtained by prompting LLMs.


\paragraph{SQuAD-Based Questions.}

We use the SQuAD (document, questions, and answers)
as training data for question generation.
The training process involves feeding the model document-answer pairs 
so that it learns to generate questions.
Despite SQuAD is not designed for event argument extraction,
this model can generate a (sometimes noisy) question given an event trigger and role.
It is worth noting that this model \emph{requires the answers} to generate the questions.
Therefore, SQuAD-based questions can only be used at training time.

While these questions are contextualized, and as we shall see it is beneficial to train with them,
we found that they are often not grounded on the event of interest or contain misunderstandings.
For example, the answer to Q4 from Figure \ref{f:motivation},
(\emph{What did Russia destroy 500 trucks with?})
is not \emph{oil} (\emph{oil} is the artifact of \emph{importing}).

\paragraph{Weak Supervision from LLMs.}

To generate more sound contextualized questions, we adopted a targeted approach.
We utilized a subset of 
RAMS~(500 random samples, 7\% from the training split) 
to prompt~(see details in Appendix~\ref{a:gpt4_question_generation}) GPT-4 with
documents, event triggers, and the corresponding argument roles. 
We instruct GPT-4 to generate five wh-questions
based on the input prompt, yielding nearly 9,000 questions asking for the arguments of events. 

We then train a T5 model to generate questions using weak supervision and the 9,000 questions~(output)
paired with the documents, event triggers and their roles~(input).
This model generates role-specific questions not only for RAMS,
but, as we shall see, also for other corpora such as WikiEvents.
For example, in Figure ~\ref{f:motivation}, 
Q5: \emph{What is being imported from ISIS-held territory in Syria and Iraq?} 
is generated for the \emph{artifact} role of the \emph{importing} event. 
Note that the questions contain other arguments of \emph{importing}
(\emph{Syria and Iraq}: \emph{origin})
as well as additional information from the document (\emph{ISI-held territory}).
Additionally, this question does not explicitly mention the role of interest (\emph{artifact}).
These questions capture both argument and non-argument cues from the document,
facilitating the generation of nuanced and semantically diverse questions.
Furthermore, using argument roles as input, as opposed to argument spans,
facilitates event-grounded knowledge transfer to other datasets.

\section{Experiments}
\label{s:experiments}
We formulate the problem of event argument extraction 
as a question-answering task to empirically assess 
the proposed question generation strategies.  
Our experiments\footnote{Code including dataset transformed into questions and answers are available at \href{https://github.com/nurakib/event-question-generation}{https://github.com/nurakib/event-question-generation}} are designed to determine
the optimal strategy as well as the
optimal combination of uncontextualized and contextualized questions. 
Additionally, we conduct zero- and few-shot inference on RAMS 
to benchmark the GPT-3 model
w.r.t our simplest question generation approach. 

\paragraph{Models.}
Template-based uncontextualized questions need no model. 
We use the GPT-4 model to generate 
prompt-based uncontextualized questions. 
For contextualized questions, we fine-tune T5~\citep{raffel2020exploring}
with either SQuAD or the weakly supervised data obtained by prompting GPT-4 (weak supervision from LLMs).
Regardless of how questions are generated,
we fine-tune BART~\cite[base and large versions]{lewis-etal-2020-bart} to answer them (i.e., find arguments for a given role of an event trigger).
We use Pytorch~\cite{paszke2019pytorch} and HuggingFace transformers~\cite{wolf-etal-2020-transformers}.
The only exceptions are GPT-3 and GPT-4, which have their own API.

\paragraph{Combining Uncontextualized and Contextualized Questions.}
In addition to using a single strategy, we combine uncontextualized and contextualized questions during training. At testing time, we use uncontextualized questions only based on our initial empirical findings (Table \ref{t:results_single}).

\paragraph{Datasets.}
We work with two datasets: 
RAMS~\cite{ebner-etal-2020-multi} 
and WikiEvents~\cite{li-etal-2021-document}.
The events and arguments in RAMS are annotated based
on the AIDA-1\footnote{\href{https://www.darpa.mil/program/active-interpretation-of-disparate-alternatives}{AIDA-1}} ontology.
Documents are sourced from news articles, resulting in 139 event types and 65 argument roles.
The event-argument structures in RAMS are annotated within a 5-sentence window;
there is one event trigger annotated per document.

WikiEvents is quite different than RAMS.
It adopts the 
KAIROS\footnote{\href{https://www.darpa.mil/program/knowledge-directed-artificial-intelligence-reasoning-over-schemas}{KAIROS}} ontology, 
sources articles from Wikipedia, and encompasses 50 event types and 59 argument roles. 
WikiEvents offers a more detailed perspective 
by annotating event-argument structures across entire documents (much longer than five sentences), 
facilitating the identification of coreferent mentions and allowing
annotation of multiple events within the same document. 

Table \ref{t:rams_stats} in Appendix \ref{a:rams} presents basic statistics of the RAMS and WikiEvents datasets.

\subsection{Zero- and Few-Shot Inference with GPT-3}
\label{ss:approaches_gpt3}
Large language models are credited with having emergent abilities~\cite{wei2022emergent}.
They are also capable of in-context learning~\cite{brown2020language},
meaning that they can (presumably) solve down stream tasks with a small set of data samples
when given instructions~\cite{wang-etal-2022-super}.
Our methodology uses them to generate questions;
these questions are used
(a)~fine-tune T5 for question generation
and
(b)~BART for event-argument extraction.
In order to test the aforementioned abilities when it comes to extracting event-argument structures across sentences,
we experiment with GPT-3 with zero- and few-shot prompts:

\noindent
\textbf{Zero-Shot.}
We prompt GPT-3 with the input document (five sentences) and the questions generated with the template-based approach~(Section~\ref{ss:generating_independent_questions}).

\noindent
\textbf{Few-Shot.}
We prompt GPT-3 with the same zero-shot prompt but preceded by two examples from the training split randomly selected
(and using the same format than the expected answer).
These examples also include questions without answers.

We present details and examples of the prompts in appendix~\ref{a:gpt3}.
We do not elaborate on them here because they obtain poor results (Section~\ref{p:results_gpt3}).

\subsection{Evaluation}
\label{ss:eval}
All our fine-tuned BART models are evaluated using the official evaluation scripts.
That is, we use P, R, and F1 after comparing (exact match) predictions and the ground truth based on token spans~(i.e., indexes of the first and last tokens, not the text in the token span).
Our rationale is that it is the only way to compare with previous work.

For zero- and few-shot inference with GPT-3, we use a more lenient evaluation.
Rather than checking for spans, we consider a prediction correct if the generated text includes the text of the expected answer.
Since there can be multiple spans in the document matching the output of GPT-3,
this evaluation is more lenient and better accommodates generative models.
Despite being more lenient, GPT-3 obtains substantially lower results.

\section{Quantitative Results and Analyses}
\label{s:results}

\begin{table}
\small
\centering
\begin{tabular}{l r@{\hspace{.1in}}r}
\toprule
 & \multicolumn{1}{c}{Base} & \multicolumn{1}{c}{Large} \\ \midrule
Uncontextualized Questions\\
~~Template-based & 48.5\scriptsize{\(\pm\)0.81}\makebox[0pt][l]{$^{}$}                 & 52.3\scriptsize{\(\pm\)1.72}\makebox[0pt][l]{$^{}$}  \\
~~Prompt-based, zero-shot              & 48.8\scriptsize{\(\pm\)0.88}\makebox[0pt][l]{$^{}$}                 & 51.8\scriptsize{\(\pm\)1.12}\makebox[0pt][l]{$^{}$}  \\
~~Prompt-based, few-shot & 47.9\scriptsize{\(\pm\)1.95}\makebox[0pt][l]{$^{}$}                 & 53.1\scriptsize{\(\pm\)0.68}\makebox[0pt][l]{$^{}$}  \\
\midrule
Contextualized Questions\\
~~SQuAD-based             & n/a                 & n/a  \\
~~Weak Supervision, LLMs             & 45.2\scriptsize{\(\pm\)1.71}\makebox[0pt][l]{$^{}$}                & 46.5\scriptsize{\(\pm\)0.91}\makebox[0pt][l]{$^{}$}   \\

\bottomrule
\end{tabular}

\caption{Results (F1, mean and standard deviation of five runs)
  obtained with the test split of RAMS using each strategy to generate questions.
  For uncontextualized questions, zero-shot prompts outperformed the few-shot prompts,
  but the opposite is true of contextualized questions.
  Contextualized questions yield substantially lower results than uncontextualized questions,
  although combining both is the best strategy (Table \ref{t:results_combined}).
  }
\label{t:results_single}
\end{table}

\begin{table}[t]
\small
\centering
\begin{tabular}{l rr}
\toprule
 & \multicolumn{1}{c}{Base} & \multicolumn{1}{c}{Large} \\ \midrule
Template-based                     & 48.5\scriptsize{\(\pm\)0.81}\makebox[0pt][l]{$^{}$}                 & 52.3\scriptsize{\(\pm\)1.72}\makebox[0pt][l]{$^{}$}\\
+ SQuAD-based               & 49.0\scriptsize{\(\pm\)0.86}\makebox[0pt][l]{$^{}$}                 & 53.1\scriptsize{\(\pm\)0.34}\makebox[0pt][l]{$^{*}$}  \\
+ Weak Supervision, LLMs                & \bf 51.5\scriptsize{\(\pm\)0.26}\makebox[0pt][l]{$^{*}$}             & \bf 54.5\scriptsize{\(\pm\)0.68}\makebox[0pt][l]{$^{*}$}  \\
\midrule
Prompt-based, zero-shot                  & 48.8\scriptsize{\(\pm\)0.88}\makebox[0pt][l]{$^{}$}                 & 51.8\scriptsize{\(\pm\)1.12}\makebox[0pt][l]{$^{}$} \\
+ SQuAD-based               & 49.6\scriptsize{\(\pm\)0.93}\makebox[0pt][l]{$^{}$}                 & 52.7\scriptsize{\(\pm\)1.87}\makebox[0pt][l]{$^{*}$}  \\
+ Weak Supervision, LLMs                & 50.3\scriptsize{\(\pm\)0.46}\makebox[0pt][l]{$^{*}$}                 & 53.4\scriptsize{\(\pm\)1.48}\makebox[0pt][l]{$^{*}$}  \\
\midrule
Prompt-based, few-shot                   & 47.9\scriptsize{\(\pm\)1.95}\makebox[0pt][l]{$^{}$}                 & 53.1\scriptsize{\(\pm\)0.68}\makebox[0pt][l]{$^{*}$} \\
+ SQuAD-based               & 49.3\scriptsize{\(\pm\)0.53}\makebox[0pt][l]{$^{}$}                 & 52.6\scriptsize{\(\pm\)0.12}\makebox[0pt][l]{$^{}$}  \\
+ Weak Supervision, LLMs                & 49.7\scriptsize{\(\pm\)0.77}\makebox[0pt][l]{$^{}$}                 & 53.5\scriptsize{\(\pm\)1.31}\makebox[0pt][l]{$^{*}$}  \\
\midrule
Previous Work (Top 5) \\
~~APE~\citeyearpar{zhang-etal-2023-overlap}                     & 51.6                 & 54.3  \\
~~APE (Single)~\citeyearpar{zhang-etal-2023-overlap}            & 49.6                 & 51.7  \\
~~SPEAE~\citeyearpar{nguyen-etal-2023-contextualized}           & 51.1                 & 53.3  \\
~~TabEAE~\citeyearpar{he-etal-2023-revisiting}                  & n/a                  & 52.7  \\
~~TARA~\citeyearpar{yang-etal-2023-amr}                         & 48.0                 & 52.5  \\
~~PAIE~\citeyearpar{ma-etal-2022-prompt}                        & 49.5                 & 52.2  \\

 \bottomrule
\end{tabular}
\caption{Results (F1, mean and standard deviation of five runs)
  obtained with the test split of RAMS using combination of uncontextualized and contextualized questions.
  The best strategy is to blend template-based (uncontextualized) and weak supervision with LLMs (contextualized) questions.
  We indicate statistically significantly better results (McNemar test \cite{McNemar1947}, $p<0.01$)
  with respect to \emph{Template-Based Questions} with an asterisk ($^{*}$).}
\label{t:results_combined}
\end{table}

\subsection{Results with RAMS}
\label{ss:rams_results}
Table \ref{t:results_single} outlines the results (F1) with our BART models (base and large)
with RAMS and using question from a single strategy.
For uncontextualized questions, 
zero-shot prompt-based questions performed best with the base model (F1: 48.8 vs. 45.2--48.5), 
while few-shot prompt-based questions excelled with the large model (F1: 53.1 vs. 46.5--52.3). 
Weakly supervised questions negatively impact performance (F1: 45.2 (base) and 46.5 (large)), 
possibly due to high variation and
the requirement of resolving multiple contextual cues. 
This lead us to recommend uncontextualized questions
when using a single strategy to generate questions.

We report the results with BART models fine-tuned with uncontextualized and contextualized questions
in Table ~\ref{t:results_combined}.
Combining template-based and weakly supervised questions 
from RAMS yields the best results for both base and large models,
showcasing a 3.0 F1 improvement (48.5 vs. 51.5) for the base model and
a 2.2 F1 improvement (52.3 vs. 54.5) for the large model. 
This approach slightly surpasses state-of-the-art models for the large variant.
We hypothesize that contextualized~(i.e., questions grounded on the event and document of interest) force the model to learn 
semantic interactions between event arguments and contextual cues 
due to the framing of the question.
Additionally, combining SQuAD-based questions~(also contextualized) at training time also 
benefit results across the board.
This is true despite their inherent noise and wording discrepancies related to events other than the event of interest (Section~\ref{ss:generating_dependent_questions}). The results in Table \ref{t:results_single} confirm the advantage of asking uncontextualized questions,
and Table \ref{t:results_combined} shows the consistent benefit of combining uncontextualized and contextualized questions.

\paragraph{Comparison with Previous Work on RAMS.}
Our best model slightly outperforms the best published results with RAMS (Table \ref{t:results_combined}, bottom block),
APE~\citep{zhang-etal-2023-overlap}, which achieves an F1 of 54.3 (ours: 54.5). 
APE relies on overlapping role annotations from 
multiple event argument extraction corpora (ACE05, RAMS, and WikiEvents). 
Notably, when APE is trained only with RAMS, it achieves 51.7 F1 (ours: 54.5), 
indicating a significant performance drop compared to other models.
~\newcite{nguyen-etal-2023-contextualized} proposed 
generating contextualized soft prompts
for event argument extraction. 
This work achieves the second-closest F1-score to our model.
Generating soft prompts depends on a set of relevant documents 
and manually initialized prompts, whereas our event-grounded 
question generation approach learns from a small portion of RAMS documents
and as we shall see, it transfers to other domains (WikiEvents, Section \ref{s:wiki_results}).

\begin{table}
\small
\centering
\begin{tabular}{l rrrr}
    \toprule
    \multirow{2}{*}{} & \multicolumn{2}{c}{Base} & \multicolumn{2}{c}{Large}\\ \cmidrule(lr){2-3} \cmidrule(lr){4-5}
     & \multicolumn{1}{c}{F1} & \multicolumn{1}{c}{\%$\Delta$F1} & \multicolumn{1}{c}{F1} & \multicolumn{1}{c}{\%$\Delta$F1} \\
    \midrule
    2 before     & 26.89      & +53.3     & 30.53    & +6.8 \\
    1 before     & 35.04      & +19.5     & 40.00    & +19.1 \\
    same         & 56.19      &  +3.7     & 59.38    & +11.5 \\
    1 after      & 24.82      & +19.3     & 27.84    &  +3.5 \\
    2 after      & 15.69      & +37.2     & 21.05    & +12.6 \\
    \bottomrule
\end{tabular}
\caption{Results with RAMS and our best model (boldfaced in Table~\ref{t:results_combined})
  broken down by distance (\# sentences) between arguments and events.
  \%$\Delta$F1 indicates the relative improvement with respect to training only with template-based questions (first supervised system in Table \ref{t:results_combined}).
  Our approach benefits all arguments, especially those that are not in the same sentence than the event.}
\label{t:results_distance}
\end{table}

\paragraph{Are Inter-Sentential Arguments Harder?}
Table \ref{t:results_distance} details the results of our best models
(combining template-based and weak supervision from LLMs, boldfaced in Table~\ref{t:results_combined})
broken down by distance (number of sentences) between the event trigger and argument.
The improvements (\%$\Delta$F1) with respect to the simplest question-answering approach
(only template-based questions; first system in Table~\ref{t:results_combined})
are substantial regardless of distance between the event trigger and argument.
For the base model, we observe a substantial 37.2\% improvement 
when the argument appears two sentences after the event trigger. 
Notably, improvements remain considerable for arguments appearing 
in sentences before (53.5\% F1, 19.5\% F1) or after (19.3\% F1) the event trigger.
For the large model, the most significant improvement of 19.1\% F1 occurs 
when arguments are in the sentence before the event trigger, followed by those two sentences after~(12.6\% F1). 
Arguments from other sentences also benefit (6.8\% and 3.5\% F1). 
In essence, our approach demonstrates robustness in extracting 
event-argument structures regardless of the distance between arguments and event triggers. 
The benefits, however, are most noticeable when arguments are not in the same sentence than the event trigger.

\begin{figure}
\includegraphics[width=\columnwidth]{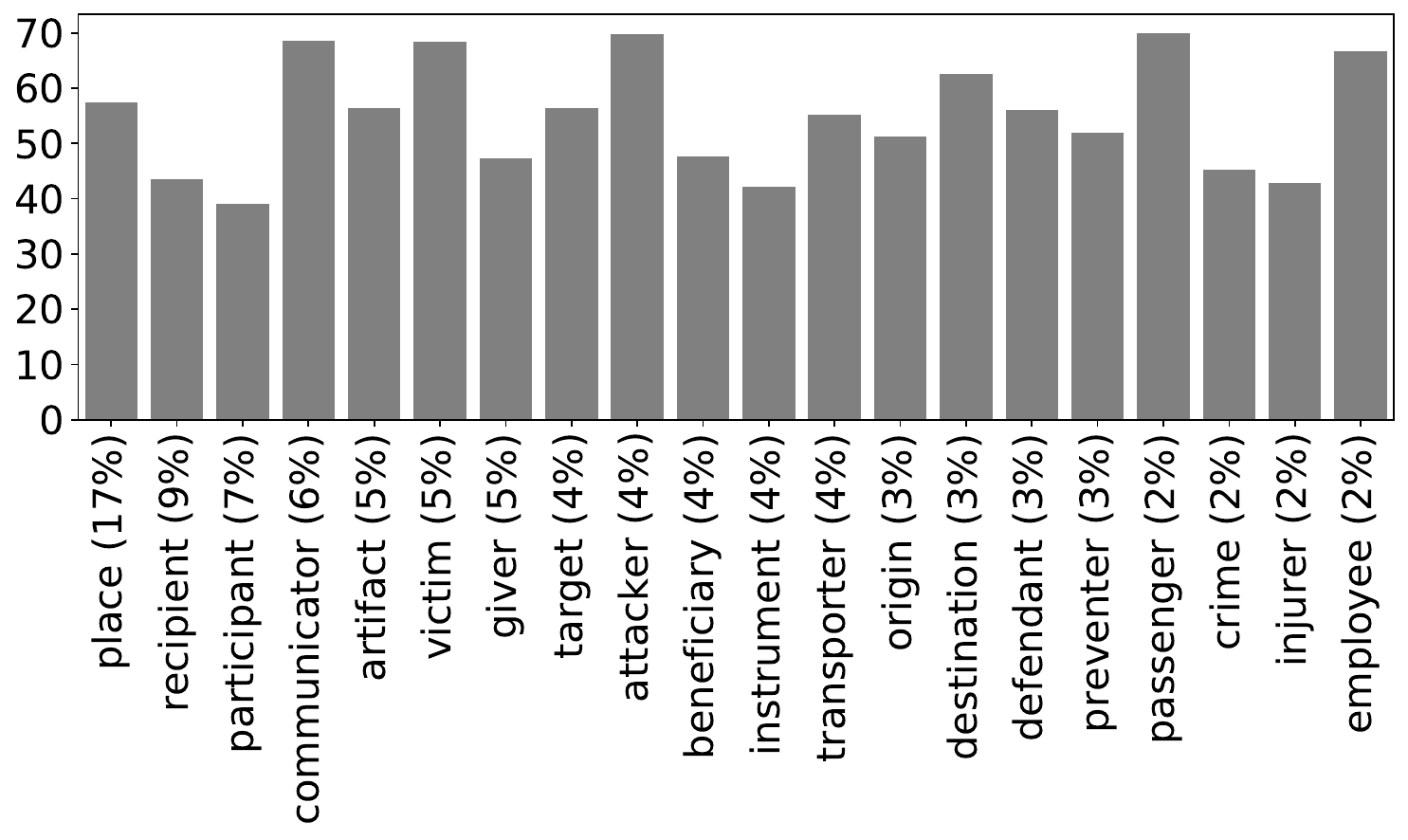}
\caption{F1 per argument of our best system (boldfaced in Table \ref{t:results_combined}).
  Frequency in training (between parenthesis) is only a weak indicator of F1.
  leading to the conclusion that some arguments are easier to learn
  (e.g., \emph{passenger} is 70\% less frequent than \emph{participant} yet the former obtains twice the F1 (0.70 vs. 0.33).}
\label{f:results_args}
\end{figure}

\begin{figure}
\includegraphics[width=\columnwidth]{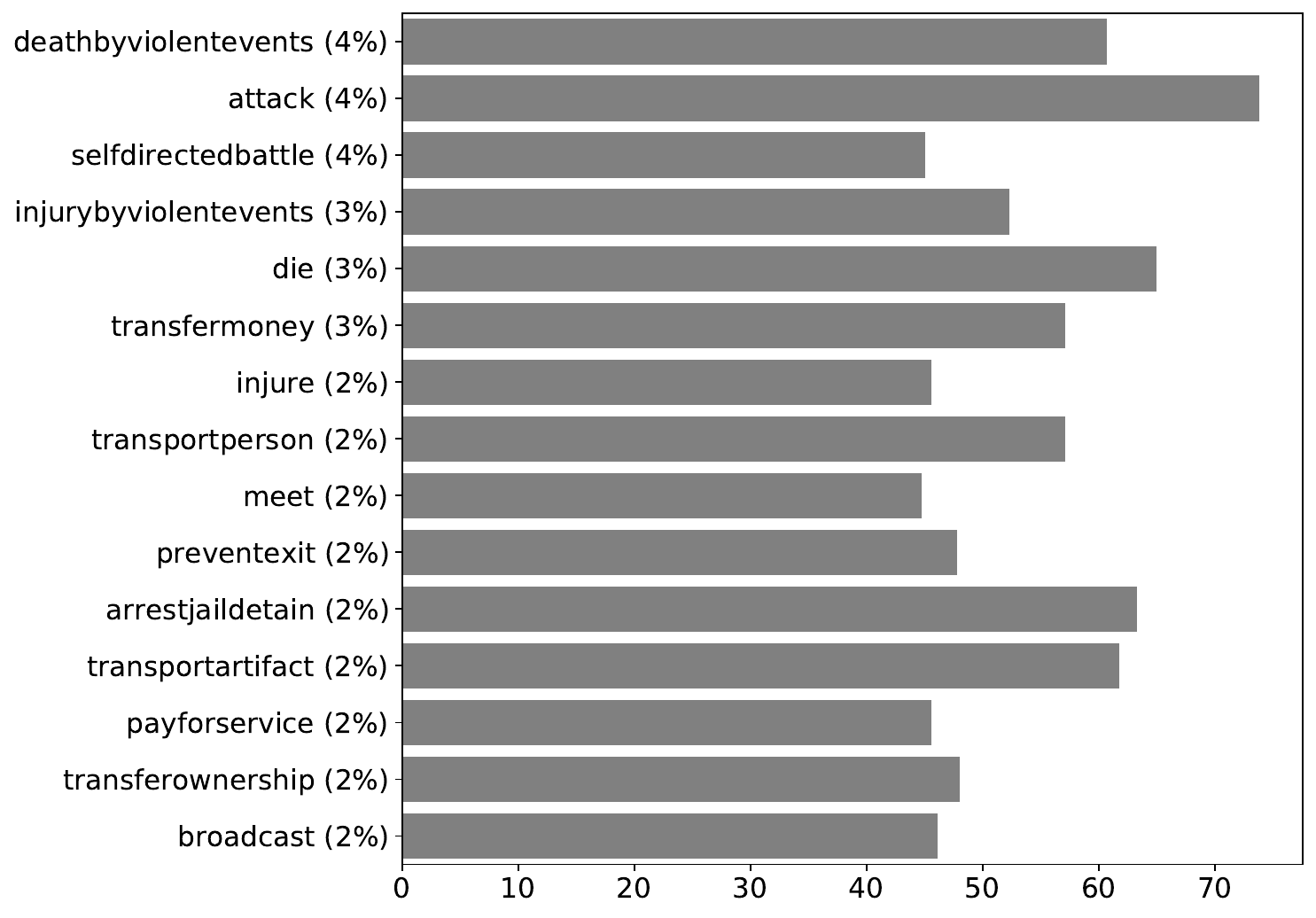}
\caption{Average F1 per event (top 15 most frequent events)
  by our best system (boldfaced in Table \ref{t:results_combined}).
  There is no clear relation between event frequency in training (between parenthesis) and F1,
  leading to the conclusion that arguments of some events are easier to learn (e.g., \emph{selfdirectedbattle} vs. \emph{payforservice})}
\label{f:results_events}
\end{figure}

\begin{figure}
\includegraphics[width=\columnwidth]{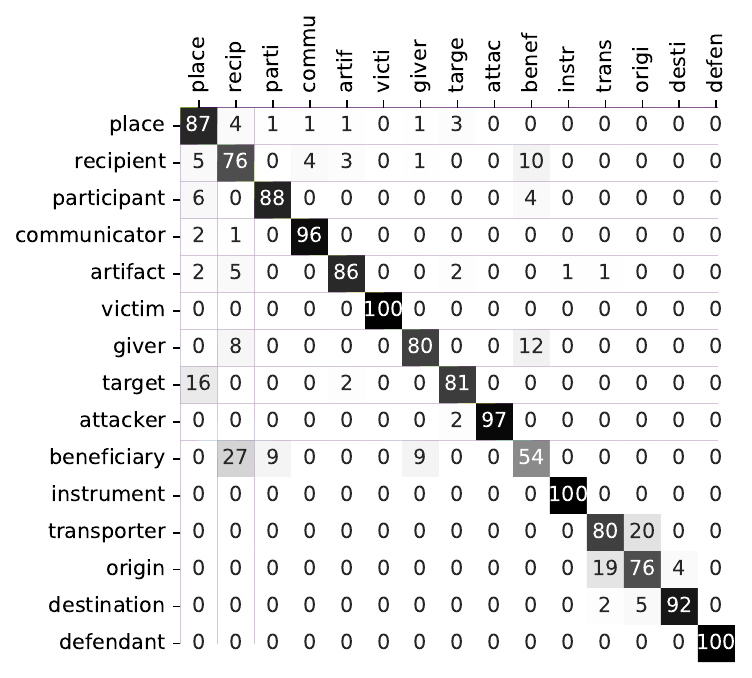}
\caption{Confusion matrix comparing gold (rows) and predicted (columns) argument roles 
  for correctly predicted argument spans (top 15 most frequent types).
  Most errors are plausible (at face value) but semantically wrong argument roles
  (e.g., mislabeling the \emph{beneficiary} as the \emph{recipient}; note that both are usually people).} 
\label{f:args_mislabeled}
\end{figure}

\paragraph{Are Frequent Arguments Easier?}
It is a common belief that the more training data the easier it is to learn.
Figure \ref{f:results_args} provides empirical evidence showing that this is not the case when predicting event-argument structures in RAMS.
We observe that per-argument F1 scores range from 0.33 to 0.70,
but there is no pattern indicating that frequency correlates with F1 score.
For example, infrequent events such as \emph{employee} and \emph{passenger}~(2\%) obtain results
as high as those obtained with communicator (6\%) and victim (5\%).


\paragraph{Are Frequent Events Easier?}
No, they are not.
Surprisingly, more training data for an event does not always lead to better results.
Figure \ref{f:results_events} shows the average F1 scores for the 15 most frequent events.
The graph shows no clear pattern between event frequency in training and F1 scores.
Indeed, arguments of events with 2\% frequency obtain F1 scores ranging from 0.33 and 0.60, a large range overlapping with the F1 scores of more frequent~events.

\paragraph{Which Arguments are Mislabeled?} 
Our best model obtains 54.5 F1.
This number is low, but the evaluation is as strict as it gets:
it expects predictions to match the exact argument span and argument role. 
Figure \ref{f:args_mislabeled} compares gold (rows) and predicted (columns) argument roles
when our best model (boldfaced in Table \ref{t:results_combined}) predicts the correct argument spans.
We observe two main trends.
First, the model mislabels arguments with labels that are (in principle) plausible but wrong given the input document.
For example, \emph{recipient}, \emph{beneficiary}, and \emph{giver} are often people but they have different semantics given an event trigger.
Second, our model mislabels arguments that could be considered correct but do not follow the RAMS annotations. 
For example, the \emph{transporter} of a \emph{transporting} event~(i.e., the person moving something)
could be the \emph{origin} or the event,
but RAMS reserves that argument role for the location where \emph{transporting} started.
We hypothesize that our model is leveraging the knowledge acquired about \emph{transporter} and \emph{origin}
prior to our training with RAMS (and it never overcame this knowledge during training).

\begin{table}
\small
\centering
\begin{tabular}{l rrr}
\toprule
                            & \multicolumn{1}{c}{P}           & \multicolumn{1}{c}{R}            & \multicolumn{1}{c}{F1} \\ \midrule

Ours, Base            & 57.1\scriptsize{\(\pm\)0.48}             & 42.1\scriptsize{\(\pm\)1.26}       & 48.5\scriptsize{\(\pm\)0.81} \\
Ours, Large           & 56.0\scriptsize{\(\pm\)1.31}             & 49.1\scriptsize{\(\pm\)2.14}       & 52.3\scriptsize{\(\pm\)1.76} \\ \midrule

\multicolumn{3}{l}{GPT3} \\
~~~Zero-shot               & 27.3                & 21.4       & 24.0 \\
~~~Few-shot                & 32.6                & 29.1       & 30.7 \\\bottomrule

\end{tabular}
\caption{Results obtained with our simplest question generation strategy (\emph{template-based}) and GPT-3 (\emph{text-davinci-003}) with zero- and few-shot prompts.
}
\label{t:gpt3_comparision}
\end{table}

\paragraph{Comparison with GPT-3.}
\label{p:results_gpt3}
Zero-shot and few-shot prompting with GPT-3 falls short in regards to performance with respect to 
our question-answering approach.\footnote{We only report one run for zero- and few-shot with GPT-3 as we do not tune it.}
The results in Table \ref{t:gpt3_comparision} show that GPT-3 obtains better results 
in a few-shot in-context learning setting compared to a zero-shot setting.
Yet, the performance lags behind our template-based question answering model.
It is important to acknowledge the potential of prompt engineering~\cite{gu-etal-2022-ppt} to improve performance. We reserve the avenue of prompt engineering for future work.

\begin{table}
\small
\centering
\begin{tabular}{l rr}
\toprule
 & \multicolumn{1}{c}{Base} & \multicolumn{1}{c}{Large} \\ \midrule
Template-Based Questions                                   & 63.6\scriptsize{\(\pm\)1.24}\makebox[0pt][l]{$^{}$}                 & 64.7\scriptsize{\(\pm\)0.76}\makebox[0pt][l]{$^{}$}  \\
+ Weak Supervision, LLMs                              & 66.3\scriptsize{\(\pm\)0.24}\makebox[0pt][l]{$^{*}$}                & 67.7\scriptsize{\(\pm\)1.13}\makebox[0pt][l]{$^{*}$}  \\

\midrule

Previous Work (Top 5) \\
~~APE~\citeyearpar{zhang-etal-2023-overlap}                     & 66.0                 & 68.7  \\
~~APE (Single)~\citeyearpar{zhang-etal-2023-overlap}            & 62.1                 & 65.8  \\
~~TabEAE~\citeyearpar{he-etal-2023-revisiting}                  & n/a                  & 66.5  \\
~~SPEAE~\citeyearpar{nguyen-etal-2023-contextualized}           & 66.2                 & 66.1  \\
~~PAIE~\citeyearpar{ma-etal-2022-prompt}                        & 63.4                 & 65.3  \\
~~AHR~\citeyearpar{ren-etal-2023-retrieve}                      & 46.1                 & 63.4  \\

 \bottomrule
\end{tabular}
\caption{Results with WikiEvents (F1, mean and standard deviation of five runs)
  using template-based questions and the best strategy with RAMS (boldfaced in Table \ref{t:results_combined}).
  We reuse the weakly supervised question generation model trained on a sample of RAMS.
  Results are statistically significantly better (McNemar test~\cite{McNemar1947}, $p<0.01$, indicated with $^{*}$).
  }
\label{t:results_wiki}
\end{table}

\begin{table*}
\small
\definecolor{goldenbrown}{rgb}{0.6, 0.4, 0.08}

\begin{tabular}{p{.75in} p{5.25in}}
\toprule
Error Type & Example \\ \midrule

Alternatives (42.17\%) & [\ldots] The military operation is the most complex carried out in {\textcolor{blue}{ \argsubg{Iraq}{predicted\_place}}} since US forces withdrew from { \textcolor{goldenbrown} {\argsubg{the country}{gold\_place}}} in 2011. Last week, the UN said it was bracing itself for the world's biggest and most complex humanitarian effort following the \argsubg{battle}{event\_trigger}, which it expects will displace up to one million people [\ldots]
\\  \addlinespace

Partial spans (33.73\%) & [\ldots] US Secretary of State Hillary Clinton who allowed this Gulf State nation to secure the 2022 World Cup — and that {\textcolor{goldenbrown}{ [the }}{\textcolor{blue} {\argsubg{Qataris}{predicted\_giver}}}{\textcolor{goldenbrown}{]\textsuperscript{\textsc{gold\_giver}}}} were so appreciative of they \argsubg{donated}{event\_trigger} millions to {\textcolor{goldenbrown}{ [the }} {\textcolor{blue} {\argsubg{Clinton Foundation.}{predicted\_beneficiary}}}{\textcolor{goldenbrown}{]\textsuperscript{\textsc{gold\_beneficiary}}}} [\ldots]
\\  \addlinespace

Wrong spans (15.66\%) & [\ldots] Mr. Assange spoke from the {\textcolor{goldenbrown}{\argsubg{Ecuadorean Embassy}{gold\_preventer}}} in London, where he has been holed up for four years. {\textcolor{blue} {\argsubg{Sweden}{predicted\_preventer}}} is seeking his extradition for an investigation into sexual misconduct allegations; his supporters have expressed fear that if he is \argsubg{arrested}{event\_trigger}, he could be sent to the United States and prosecuted for publishing leaked documents. [\ldots] 
\\ \addlinespace

Two or more arguments (8.04\%) & [\ldots] On Monday, the { \textcolor{goldenbrown}{\argsubg{U.S.}{gold\_participant\_2}}} and { \textcolor{goldenbrown} {\argsubg{Russia}{gold\_participant\_1}}} {\textcolor{blue}{ $\textsuperscript{predicted\_participant}$}} entered a The \argsubg{ceasefire}{event\_trigger} agreement in Syria, in part to get humanitarian aid to cities like Aleppo. The ceasefire does not apply to strikes targeting ISIS. [\ldots] \\

\bottomrule

\end{tabular}

\caption{Most common errors made by our best performing model on RAMS
(\emph{combining template-based questions with questions obtained with a weakly supervised model}, boldface in Table \ref{t:results_combined}).}
\label{t:errors}
\end{table*}

\subsection{Results with WikiEvents}
\label{s:wiki_results}
We also evaluate the effectiveness of our proposed 
question generation strategy (best performing approach with RAMS, boldfaced in Table \ref{t:results_combined}) 
on the WikiEvents dataset.
To formulate uncontextualized questions, 
we employed a template-based approach utilizing the event ontology provided with the dataset.
We reuse the question generation models used with RAMS,
including the weakly supervised model trained with questions obtained via prompting with documents, event triggers and arguments from RAMS.
In other words, there is no WikiEvents-specific component or fine-tuning.

The results (Table~\ref{t:results_wiki}) demonstrate 
that our approach transfers across corpora---recall that WikiEvents contains documents in different domains than RAMS as well as different events and argument roles.
Indeed, the best strategy to combine uncontextualized and contextualized questions yields statistically significantly higher results.
While we do not outperform all state-of-the-art models, 
particularly for larger models, our results remain competitive despite the question generation component was trained with RAMS.

\section{Qualitative Error Analysis}
\label{s:error_analysis}

We close our analyses by examining the errors made by the best model (boldfaced in Table \ref{t:results_combined}).
We discuss linguistic commonalities in either the input documents or model predictions
observed in a manual analysis of 83 errors from 100 documents.
The most common error type (42.17\%, Table \ref{t:errors}) is predicting \textit{alternative spans}. In the example, our model predicts a coreferent mention that is counted as an error. 
This allowed us to raise the issue of rethinking 
the evaluation of
event argument extraction; only one gold span matching may not be ideal.
The second most common error type (33.73\%, Table \ref{t:errors}) is predicting \textit{partial spans} (either shorter or longer than the gold).
The differences include articles, conjunctions, numbers, and detailed descriptions complementing entities.
Completely \textit{wrong spans} are much less likely (15.66\%).
Our model is limited to predicting one span per argument role,
thus it will always make errors with events that have two instances of the same argument role (8.04\% of errors).
A previously reported by~\newcite{zhang-etal-2020-two},
we found that some model errors (7.3\%) appear to be due to annotation errors---no annotations are perfect, and RAMS is not an exception.
For example, the test set includes the following: 
\textit{he raised the \argsubg{funds}{recipient} privately and \argsubg{reimbursed}{event\_trigger} the city [\ldots]}.

\section{Weakly Generated Question Analysis}
\label{s:question_analysis}

\begin{figure*}[h]
\centering
\input{figs/question_analysis_tmp}
\captionsetup{width=\textwidth}
\caption{Examples of three categories of questions generated by the weakly supervised T5 model. 
Text highlighted in green indicates the events, while red text indicates the arguments within the RAMS documents. Argument roles are mentioned in parenthesis. \emph{Question about the event and the argument role} are grounded on both the event and the argument role, and \emph{Question about the event} are only grounded on the event. \emph{Question about neither the event or the argument role} are irrelevant to the event and the argument role. \emph{Generated Question} are the outputs generated by the weakly supervised model; and \emph{Expected Question} refers to human-written questions exemplifying perfect questions~(hypothetical) we would like to generate.}
\label{f:q_analysis}
\end{figure*}

Our empirical assessment across two corpora for event argument extraction indicates that training with weakly supervised questions yields positive results.
To further understand this, we conduct a manual analysis of 100 random questions 
generated by the weakly supervised model. 
This analysis evaluates the generated questions
for their relevance and accuracy 
in relation to the events and argument roles detailed within the document.
Overall, we found three different categories after analyzing 100 questions.
The first category of questions are targeting both the event and the argument role, 
while the second category of questions focuses solely on the event. 
The third category of questions lacks relevance to both the event and the argument role. 
Figure \ref{f:q_analysis} exemplifies the 3 categories using the output from the weakly supervised T5 model.
We observe that 73\% questions generated  
are grounded in both the event and the argument role (\emph{first category}). 
Another 23\% questions are grounded in the event of interest (\emph{second category}) but aim to extract different information than the role of interest. Only 4\% questions are grounded in neither the event nor argument role of interest (\emph{third category}).

\section{Conclusions}
\label{s:conclusions}

We have presented several question generation strategies for asking questions
about the arguments of an event.
Our approach can be used seamlessly with any corpora;
the only requirement is a list of events and the arguments they may take, information that existing corpora come with.
Combining uncontextualized and contextualized questions obtains the best results.
Contextualized questions incorporate cues about other arguments of the event of interest
as well as other parts of the document,
providing our BART-based models with information that is not otherwise available.
Crucially, our strategies to generate questions do not require manual intervention.
Additionally, the weakly supervised models to generate questions transfer across corpora (from RAMS to WikiEvents; different corpora, events, and arguments roles).

\clearpage
\section*{Limitations}
This work presented in this paper has several limitations.
Our weakly supervised approach brings certain advantages in event argument extraction.
This have been assessed through empirical analysis. 
While we have observed the effectiveness of incorporating
questions during the training phase, 
we have not reported an in-depth analysis into
why this method is successful across all samples. 
It remains unclear whether the addition of all 
contextualized questions is universally beneficial.
To address this, employing an adaptive selection strategy
could prove valuable in determining the instances 
where incorporating questions is most advantageous.

Our model predicts at most one argument per event trigger. 
Thus, any event trigger that has two arguments with the same role is 
guaranteed to yield an error (Section \ref{s:error_analysis}). 
Addressing this limitation would necessitate further research, 
particularly in the context of multi-turn question answering, 
to enhance the model's capabilities.

We compare our fine-tuned model and GPT-3 with template-based questions.
It can be argued that there might be better alternative way of prompting the GPT-3 model.

\section*{Acknowledgments}
This research was supported by a grant from the U.S. Office of Naval Research (N00014-22-1-2596).
We would like to thank the anonymous reviewers for their insightful comments and suggestions.

\section*{Ethics Statement}
The authors state that this work is in accordance with the ACL Code of Ethics and does not raise ethical issues.


\bibliography{refs}
\bibliographystyle{naacl2024-latex/acl_natbib}

\appendix

\section{QA Model and Hyperparameters}
\label{a:hyperparameters}

In this section, we provide an overview of the model and hyperparameters used for finetuning question answering models. We leverage the BART model to generate representations for the question and document pairs. By feeding the question and the document as input to the BART model, we obtained the contextualized representation of the combined text. Then, we employ a task-specific layer that operates on top of these representations. This layer is responsible for predicting the start and end offsets of the argument spans.

The output layer is trained using the cross-entropy loss function to minimize the discrepancy between the predicted offsets and the ground truth offsets. Initially, we train all the models using the same random seed, 42. We use five different learning rates [1e-5, 2e-5, 3e-5, 4e-5, 5e-5] and dropout values [0.1, 0.2, 0.3] to tune the performance of the models. In order to determine the optimal hyperparameters, we evaluate the models' performance on the validation dataset. This allows us to select the final hyperparameters that yield the best result. Based on the optimized values for learning rate and dropout, we retrain the model on four new random seeds (4, 13, 52, 57) and report the F1-scores based on the average and standard deviation of the five runs. We list all the hyperparameters in Table \ref{t:parameters}.  

To mitigate the risk of overfitting and ensure efficient training, we incorporated the technique of early stopping. If the loss function fails to show  improvement over 10 consecutive epochs, training is terminated before completing 50 epochs. 

\begin{table}
\centering
\begin{tabular}{l c}
\toprule 
Name                            & Value           \\ \midrule

learning rates                  &  1e-5, 2e-5, 3e-5, 4e-5, 5e-5          \\ 
epochs                          &  50          \\ 
patience 			     		&  10   		\\
dropout                         &  0.1, 0.2, 0.3 \\
training batch                  &  8       \\
validation batch                &  4         \\
test batch                      &  4          \\
max length                      &  512          \\
loss function                   &  cross-entropy  \\
optimizer                       &  Adam          \\ \bottomrule

\end{tabular}
\caption{Hyperparameters to finetune the question answering model.}
\label{t:parameters}
\end{table}
\section{Datasets Statistics}
\label{a:rams}

\begin{table}[h!]
\small
\centering
\begin{tabular}{llrrr}
\toprule 
Dataset                       & Split           & \#Docs        & \#Events      & \#Args \\ \midrule

					          & Train			& 3194			 & 7329           & 17026 \\
RAMS					      & Dev			    & 399			 & 924 		      & 2188 \\
					          & Test			& 400			 & 871 		      & 2023 \\ 
\midrule
							  & Train			& 206			 & 3241 		  & 4542 \\
WikiEvents					  & Dev			    & 20			 & 345 		      & 428 \\
							  & Test			& 20			 & 365 		      & 566 \\

\bottomrule

\end{tabular}

\caption{Statistics of the RAMS and WikiEvents dataset.}
\label{t:rams_stats}
\end{table}
 
RAMS and WikiEvents datasets share the common objective of annotating events 
and their associated arguments within a document. 
RAMS is licensed under \emph{Apache License 2.0} and WikiEvents is licensed under \emph{MIT License}.
Table \ref{t:rams_stats} presents basic statistics of the RAMS and WikiEvents datasets.
\section{Question Generation Prompts}
\label{a:gpt4_question_generation}
We utilize the GPT-4 model to create event argument role specific questions in two scenarios:
uncontextualized and contextualized. 
In the uncontextualized scenario, we generate one question for each argument role. 
For contextualized questions, we prompt the model using 500 samples from RAMS 
and train a T5-base model to generate questions for the remaining data samples.

\begin{figure}[t!]
\centering
\input{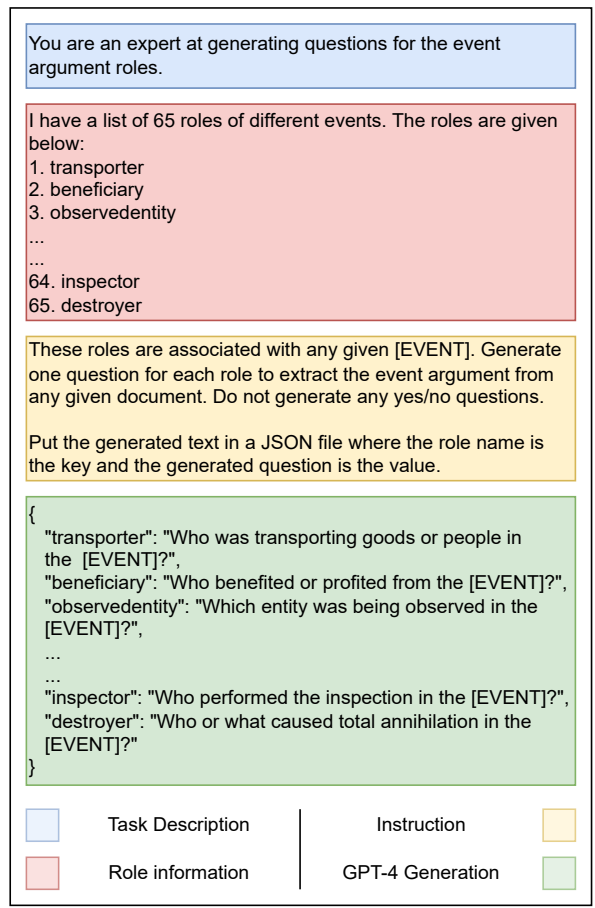}
\caption{Example of a zero-prompt question generation. \emph{Task Description}, \emph{Role information}, \emph{Instruction} are all together considered the prompt (\emph{input}) and \emph{GPT-4 Generation} is the model output.}
\label{f:zero_prompt_question}
\end{figure}

\paragraph{Uncontextualized Question Generation Prompts.}
We generate uncontextualized questions 
using the the GPT-4 model in OpenAI playground.\footnote{\href{https://platform.openai.com/playground}{OpenAI Playground.}}
Initially, we prompt the model with only argument role information, 
which we term as zero-shot prompt-based questions. 
Figure \ref{f:zero_prompt_question} presents 
the zero-shot prompt-based question generation.
We start with a task description, 
giving the model an overview of the task at hand. 
The all the argument roles are included; 
followed by task specific instruction. 
We instruct the model to produce only one question per role
and to avoid generating yes/no questions. 
Finally, we conclude the instructions 
by asking the model to format the generated questions in JSON.

\begin{figure}[t!]
\centering
\input{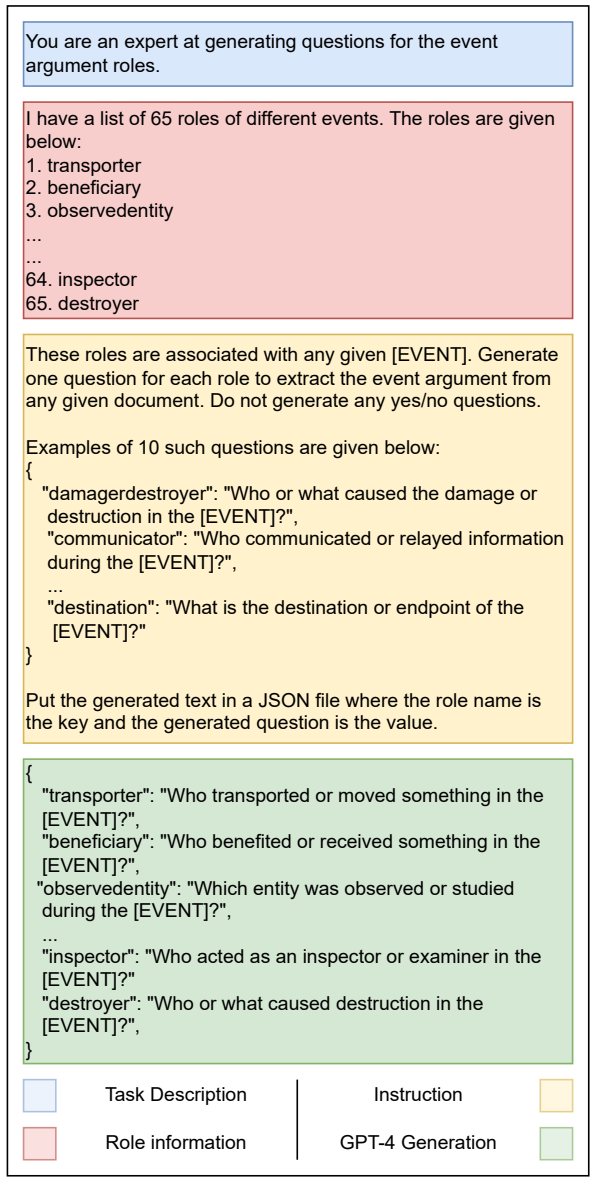}
\caption{Example of a few-prompt question generation. We added some role-specific question samples in the \emph{Instruction}. \emph{Task Description}, \emph{Role information}, \emph{Instruction} are all together considered the prompt (\emph{input}) and \emph{GPT-4 Generation} is the model output.}
\label{f:few_prompt_question}
\end{figure}

For few-shot prompt-based questions generation, 
we add ten samples from zero-shot prompt-based questions in the prompt.
Figure \ref{f:few_prompt_question} presents 
the zero-shot prompt-based question generation.

\paragraph{Contextualized Question Generation Prompt.}
We prompt the GPT-4 model to collect around 9000 event-grounded question samples.
We begin with a task description, outlining the information required for generating event-grounded questions. Next, we provide the event document from RAMS along with the task instruction. Specifically, we instruct the model to produce five distinct questions for a given event and an argument role. Although the actual event argument span is highlighted in the document, our instruction is focused on the event and argument role. The model then generates five different event-grounded questions as instructed.

\begin{figure}[t!]
\centering
\input{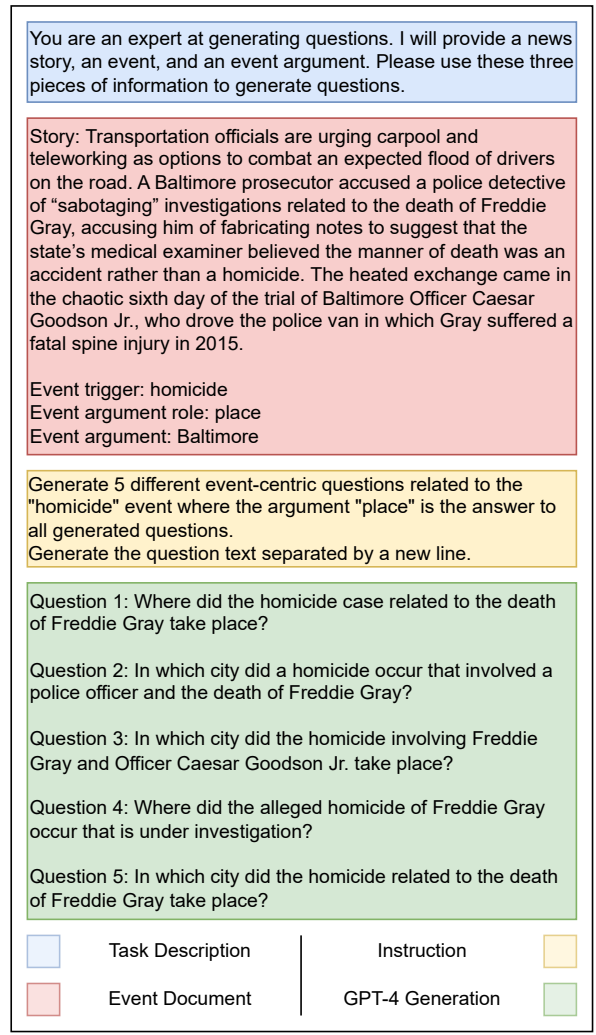}
\caption{Example of a contextualized question generation using GPT-4. \emph{Task Description}, \emph{Event Document}, \emph{Instruction} are all together considered the prompt (\emph{input}) and \emph{GPT-4 Generation} is the model output.}
\label{f:context_dependent_prompt}
\end{figure}
\section{Zero- and Few-Shot Inference Prompts}
\label{a:gpt3}

To conduct our experiments with GPT-3 in zero and few-shot settings,
we utilized the OpenAI API.
Specifically, we use the `text-davinci-003' model and the `Completion' endpoint provided by the API. To ensure consistency with the inputs used in the supervised model, we designed the prompts for the GPT-3 model in a similar manner. However, there were slight differences in how the prompts were handled between the supervised settings and the zero and few-shot experiments. In the supervised settings, we posed one question per iteration, whereas, for the GPT-3 zero-shot and few-shot experiments, we included all the questions simultaneously. We conducted a small-scale study using a subset of samples from the RAMS dataset to validate the impact of asking all questions at once compared to asking one question per iteration. Our study did not reveal any difference between the two approaches. Hence, we proceeded with asking all questions at once for the zero and few-shot experiments. This streamlined the experimentation process and also helped to reduce the costs of querying the API. 

\paragraph{Example of Zero-Shot Inference Prompt.}

In the zero-shot setting, our objective is to extract event arguments without any training examples. To accomplish this, we construct prompts with template-based questions. The GPT-3 model generates answers to these questions, which are then mapped back to the provided document to extract matched event argument spans. Figure \ref{f:zeroshot} presents a snapshot of a zero-shot prompt. 

\begin{figure}[ht!]
\centering
\input{figs/zeroshot_tmp}
\caption{Example of a zero-shot GPT-3 prompt. \emph{Test sample}, \emph{Instruction} and \emph{Asking Questions} are all together considered the prompt (\emph{input}) and \emph{GPT-3 Generation} is the model output.}
\label{f:zeroshot}
\end{figure}

\paragraph{Example of Few-Shot Inference Prompt.}

In the few-shot setting, we leverage a limited amount of training data. We randomly select two training samples from the RAMS dataset. These examples are formatted to match the inputs used during supervised training. By incorporating two training samples, we enhance the model's ability to capture event arguments and generate accurate responses. Figure \ref{f:fewshot} presents a snapshot of a few-shot prompt. 

\begin{figure}[ht!]
\centering
\input{figs/fewshot_tmp}

\caption{Example of a few-shot GPT-3 prompt. \emph{Training samples}, \emph{Test sample}, \emph{Instruction} and \emph{Asking Questions} are all together considered the prompt (\emph{input}) and \emph{GPT-3 Generation} is the model output.}
\label{f:fewshot}
\end{figure}

\end{document}